\documentclass[11pt]{article}
\usepackage[margin=1.25in]{geometry}
\usepackage{times}
\usepackage[numbers,sort&compress]{natbib}
\usepackage{hyperref}
\usepackage{url}
\usepackage{amsmath,amssymb}
\usepackage{booktabs}
\usepackage{microtype}
\usepackage{listings}
\usepackage{parskip}

\title{\textbf{Scaling Neural Network Verification with\\
Tensor Parallelism and Fully Sharded Data Parallelism}}

\author{%
  Sergei Vorobyov \\
  Lomonosov Moscow State University \\
  \texttt{sergei.vorobyov01@gmail.com}
  \and
  Eugene Ilyushin \\
  Lomonosov Moscow State University,\\
  Central University\\
  \texttt{eugene.ilyushin@gmail.com}
}

\date{}

\begin{document}

\maketitle

% -----------------------------------------------------------------------
\begin{abstract}
Formal neural network verification---proving that a network satisfies safety
properties for \emph{all} inputs in a specified domain---is bounded in practice
by GPU memory: standard implementations of bound-propagation algorithms
(IBP, CROWN, $\alpha$-CROWN) require weight and relaxation-coefficient matrices
to reside entirely on one accelerator. We adapt two parallelism techniques
originally developed for large-scale model training to the
\texttt{auto\_LiRPA}\,/\,$\alpha,\beta$-CROWN verification framework.
\textbf{Tensor Parallelism (TP)} shards both weight and $A$-matrices across GPUs,
achieving ${\approx}2\times$ peak-memory reduction at $P{=}2$; soundness is
confirmed on VNN-COMP 2022 MNIST-FC benchmarks, though bound tightness degrades
with the number of sharded zones due to forced IBP substitution for intermediate
bounds inside sharded zones. \textbf{Fully Sharded Data Parallelism (FSDP)}
shards only weight matrices with a per-layer \texttt{AllGather}, producing
bounds that are \emph{bitwise identical} to the single-GPU baseline: baseline
memory drops by 80--90\%, peak memory by 34--39\% on wide MLPs. FSDP integrates
cleanly with complete verification ($\beta$-CROWN + Branch-and-Bound) and with
convolutional layers (\texttt{BoundConv}); a complete \emph{unsat} result is
obtained for CIFAR-100 ResNet-large (VNN-COMP 2024) under FSDP. Across all
experiments the memory bottleneck in $\alpha$-CROWN+BaB mode proves to be
per-neuron alpha tensors, not weight matrices, pointing to the key direction
for future work.
\end{abstract}

% -----------------------------------------------------------------------
\section{Introduction}
\label{sec:intro}

Neural networks are increasingly deployed in safety-critical contexts where
errors are intolerable: autonomous vehicle control systems, avionics software,
and medical diagnostic algorithms. Standard testing methods are fundamentally
insufficient in such settings: a test verifies a specific input vector, not
the entire set of admissible inputs \citep{Guidotti2019Enhancing}. Formal
verification addresses exactly this gap---it provides a mathematical proof
that the network behaves correctly for \emph{every} input in a specified
region.

\subsection{Formal Verification Problem}

Let a neural network define a function $f: \mathbb{R}^n \to \mathbb{R}^m$.
The verification problem is to decide whether, for all inputs in an admissible
set $\mathcal{X} \subseteq \mathbb{R}^n$, an output property $\psi$ holds:
\[
  \forall\, \mathbf{x} \in \mathcal{X},\quad \psi(f(\mathbf{x})).
\]
In practice, verifiers solve the dual problem---they search for a
\emph{counterexample} $\mathbf{x} \in \mathcal{X}$ violating $\psi$:
\[
  \exists\, \mathbf{x} \in \mathcal{X} : \neg\psi(f(\mathbf{x})).
\]
If the query is unsatisfiable (UNSAT), the property is proved; if satisfiable
(SAT), a counterexample is returned \citep{marzari2025advancingneuralnetworkverification}.

The most widely studied property is \emph{local robustness}: for a sample
$\mathbf{x}_0$ of correct class $c$, the classifier must assign class $c$ to
all inputs within an $\varepsilon$-ball,
$\|\mathbf{x}-\mathbf{x}_0\|_p \leq \varepsilon$ \citep{DBLP:journals/corr/abs-2002-03339}.
Beyond image classifiers, functional specifications over physical variables are
used in control tasks such as ACAS~Xu \citep{Demarchi2023SupportingStandardization}.

\subsection{The VNN-LIB Standard and VNN-COMP}

Prior to 2020, each verifier used its own format for network and property
descriptions. The VNN-LIB standard \citep{VNNLIBWebsite} unified the
specification language based on SMT-LIB: input/output variables and linear
constraints over them with UNSAT/SAT semantics. Together with ONNX for weight
storage, VNN-LIB forms the infrastructure for the annual VNN-COMP competition
\citep{arXiv2212VNNCOMP2022, arXiv2312VNNCOMP2023, arXiv2412VNNCOMP2024},
where benchmarks and tools are developed independently.

\subsection{Computational Complexity}

The seminal Reluplex paper~\citep{ReluplexSemanticScholar} established that
robustness verification of ReLU networks is NP-complete via a polynomial
reduction from 3-SAT. For smooth activations (sigmoid, tanh, GELU) the
problem involves transcendental constraints and is therefore at least as hard
as the ReLU case; precise classification within the polynomial hierarchy
(or its real-arithmetic analogues such as $\exists\mathbb{R}$) remains an
active research direction~\citep{NeurIPS2021ExistsRComplete}.\footnote{The
cited \citet{NeurIPS2021ExistsRComplete} establishes $\exists\mathbb{R}$
hardness for the \emph{training} problem with smooth activations; it does not
directly settle the verification problem, but illustrates that real-arithmetic
hardness arises naturally for these activation classes.} In practice this
means a \emph{complete} verifier must use worst-case exponential
Branch-and-Bound, while an \emph{incomplete} verifier replaces it with a
polynomial relaxation at the cost of losing completeness guarantees.

\subsection{Memory Bottleneck}

The dominant practical approach to verification is \emph{bound propagation}.
$\alpha,\beta$-CROWN \citep{AlphaBetaCrownVNNCOMP23}---winner of VNN-COMP
2021--2024---runs entirely on GPU and builds linear over-approximations of the
network output set. Its bottleneck is memory: weight matrices and
linear-relaxation $A$-matrices must fit on a single accelerator. A single
hidden layer of width $h{=}262\,144$ with input dimension $4\,096$ already
requires ${\approx}4$~GB in float32 for the weight alone---and that is before
batching, $A$-matrix construction, or $\alpha$ tensors are accounted for.

This paper investigates how \emph{parameter-sharding} strategies developed for
large-scale model training can remove that constraint, and experimentally
characterises the accuracy--memory trade-offs each approach entails.

\subsection*{Contributions}
\begin{enumerate}
\item We implement \textbf{Tensor Parallelism} for \texttt{auto\_LiRPA} via
  custom \texttt{BoundLinearTP\_Col/Row} operators, achieve
  ${\approx}2\times$ peak-memory reduction at $P{=}2$, and formally analyse
  the bound-tightness degradation caused by IBP substitution for intermediate
  bounds inside sharded zones.
\item We implement \textbf{FSDP} for bound propagation in ${\approx}100$ lines,
  producing bitwise-identical bounds with 34--39\% peak and 80--90\% baseline
  memory savings on wide MLPs.
\item We integrate FSDP with \textbf{complete verification} ($\beta$-CROWN+BaB),
  introduce a \texttt{force\_synchronous} flag for fair comparison, and extend
  sharding to \textbf{convolutional layers} (\texttt{BoundConv}).
\item We identify \textbf{per-neuron alpha tensors}---not weights---as the true
  memory bottleneck in BaB mode, shifting the focus for future multi-GPU
  verification research.
\end{enumerate}

% -----------------------------------------------------------------------
\section{Verification Algorithms}
\label{sec:verifiers}

\subsection{Hierarchy of Bound-Propagation Methods}

Bound-propagation algorithms compute, for each network output, an interval
guaranteed to contain the true value for any admissible input. Three levels of
precision and cost are distinguished.

\textbf{IBP (Interval Bound Propagation)} \citep{CROWNPaper} propagates
intervals $[\mathbf{l}, \mathbf{u}]$ layer by layer. For a linear layer
$\mathbf{y} = W\mathbf{x} + \mathbf{b}$:
\[
  \mathbf{l}_y = W^+\mathbf{l}_x + W^-\mathbf{u}_x + \mathbf{b},\qquad
  \mathbf{u}_y = W^+\mathbf{u}_x + W^-\mathbf{l}_x + \mathbf{b},
\]
where $W^+ = \max(W, 0)$, $W^- = \min(W, 0)$ elementwise. IBP is cheap
($O(n)$ per layer) but over-approximates rapidly with depth.

\textbf{CROWN} \citep{CROWNPaper} computes tight linear bounds via a backward
pass analogous to backpropagation. For a ReLU neuron with pre-activation
$x \in [l, u]$, $l < 0 < u$, a linear relaxation is constructed:
$\lambda_l\, x \leq \mathrm{ReLU}(x) \leq \tfrac{u}{u-l}(x - l)$, with the
lower-bound slope $\lambda_l \in [0, 1]$. The backward pass links the network
output to the input through a chain of layer-wise linear inequalities. Writing
the resulting bound symbolically:
\begin{equation}
  \ell(\mathbf{x}) \;=\; A^{(L)}\mathbf{x} + \mathbf{b}^{(L)}
  \;\leq\; f(\mathbf{x}) \;\leq\;
  A^{(U)}\mathbf{x} + \mathbf{b}^{(U)} \;=\; u(\mathbf{x}),
  \label{eq:crown}
\end{equation}
where $A^{(L)}$ and $A^{(U)}$ are obtained by composing layer weights with
diagonal slope matrices $\Lambda_i$ (different slope choices are used for
the lower- and upper-bound branches). The key property is that
$A^{(L)}, A^{(U)}$ are \emph{global} linear functions of the original input
$\mathbf{x}$, preserving cross-variable dependencies across all layers---which
makes CROWN substantially tighter than IBP. For brevity, in the remainder of
the paper we use $A = \Lambda_L W_L \cdots \Lambda_1 W_1$ as a shorthand for
either branch of (\ref{eq:crown}).

\textbf{$\alpha$-CROWN} \citep{AlphaBetaCrownVNNCOMP23} promotes the fixed
slopes $\lambda_i$ to per-neuron optimisable parameters $\alpha_i \in [0,1]$.
Gradient ascent over $\alpha$ tightens the lower bound (or lowers the upper
bound) without any branching, yielding the tightest incomplete bounds possible
within the LiRPA framework.

\subsubsection{Incomplete vs.\ Complete Verification}

IBP, CROWN, and $\alpha$-CROWN are \emph{incomplete} (but sound) methods:
they construct an over-approximation (lower bounds never exceed true minima;
upper bounds never fall below true maxima). When the resulting bounds are
tight enough to confirm the property, verification succeeds; otherwise the
outcome is \emph{unknown}. The verifier never returns ``verified'' for a
property that does not hold.

\emph{Complete verification} \citep{betaCROWN2021} guarantees a definitive
answer via \textbf{Branch-and-Bound (BaB)}: when bounds are insufficient, the
input space is split into sub-regions (e.g.\ by fixing a ReLU state
$x \geq 0$ or $x < 0$), and bounds are recomputed for each sub-region. The
process repeats recursively until an answer is obtained or a timeout is reached.

\subsection{Alpha-Beta-CROWN}

$\alpha,\beta$-CROWN \citep{AlphaBetaCrownVNNCOMP23} unifies all methods above.
For incomplete verification it uses $\alpha$-CROWN; when that does not yield
tight enough bounds it invokes BaB. The key innovation of $\beta$-CROWN
\citep{betaCROWN2021} is encoding ReLU-split constraints directly into the
bound-propagation pass via Lagrangian dual variables $\beta$, avoiding the need
to solve a linear program at each BaB leaf. As a result, bounds for thousands
of sub-problems are evaluated in parallel on GPU as batched matrix
multiplications, and $\alpha$/$\beta$ parameters are jointly optimised by
gradient descent. The cutting-plane extension BICCOS
\citep{NeurIPS2024BnbCuttingPlanes} generates constraints from logical
implications in the BaB tree, tightening root-node bounds and reducing
branching depth.

\subsection{Other Verifiers}

\textbf{VeriNet} uses Symbolic Interval Propagation (SIP): instead of linear
inequalities, dependencies are maintained as symbolic expressions throughout the
network. It is particularly effective for input splitting---a strategy suited to
low-dimensional control tasks such as ACAS~Xu \citep{IntervalReachabilityAnalysis}.

\textbf{nnenum} \citep{NNEnumPaper} (winner of VNN-COMP 2020) dynamically
switches between levels of set-representation precision---from fast but coarse
zonotopes to accurate but expensive ImageStars---depending on current sub-problem
complexity.

\textbf{Marabou 2.0} \citep{arXiv2401Marabou2} transforms the combinatorial
search into continuous optimisation via the DeepSoI procedure (Sum of
Infeasibilities), a continuous measure of total ReLU-constraint violation that
can be minimised by gradient methods.

\textbf{NeuralSAT} \citep{NeuralSATRepo} adapts CDCL (Conflict-Driven Clause
Learning) from SAT solving: infeasible neuron-state combinations are recorded as
nogoods, pruning entire subtrees in subsequent search.

\subsection{The \texttt{auto\_LiRPA} Library}

All experiments in this work are built on \texttt{auto\_LiRPA}
\citep{AutoLiRPARepo}, the library underlying $\alpha,\beta$-CROWN.
Bound-propagation operations are described as nodes of a PyTorch computational
graph; the library automatically derives upper and lower bounds for any
intermediate node of an arbitrary architecture (MLP, CNN, Transformer).

Critically, \texttt{auto\_LiRPA} supports \emph{batch BaB}: instead of
processing BaB branches sequentially, it batches up to hundreds of thousands
of unsolved sub-problems and computes bounds for all of them in a single GPU
pass. This makes any modification to the library non-trivial: it must preserve
batched processing and remain compatible with the PyTorch computational graph.

% -----------------------------------------------------------------------
\section{Parallel Bound Propagation}
\label{sec:method}

\subsection{Parallelism Primitives}
\label{sec:primitives}

We briefly define the two strategies we adapt; readers familiar with
distributed training may skip to Section~\ref{sec:why}.

\textbf{Tensor Parallelism (TP)} \citep{MegatronLM2020} splits matrix
multiplications \emph{within} a single layer across $P$ GPUs. We adopt the
PyTorch convention in which an \texttt{nn.Linear} weight has shape
$(d_\text{out}, d_\text{in})$ and the layer computes
$\mathbf{y} = W\mathbf{x} + \mathbf{b}$. In the \emph{column-parallel} variant
the weight is split along the output dimension into shards
$W^{(r)} \in \mathbb{R}^{d_\text{out}/P \times d_\text{in}}$; each GPU $r$
computes a partial output $\mathbf{y}^{(r)} = W^{(r)}\mathbf{x}$ from a
replicated input. In the \emph{row-parallel} variant the weight is split along
the input dimension; each GPU multiplies its shard against its portion of the
input and partial results are summed via \texttt{AllReduce}:
$\mathbf{y} = \sum_r W^{(r)}\mathbf{x}^{(r)}$. A Column--Row pair requires
exactly one \texttt{AllReduce} at the Row output.

\textbf{Fully Sharded Data Parallelism (FSDP)} \citep{ZeROPaper} shards each
parameter tensor equally across $P$ GPUs. Before executing a layer, an
\texttt{AllGather} assembles the full parameter from shards; immediately after
the layer finishes, the full tensor is freed and only the shard is retained.
The net effect: at most one full weight matrix is in GPU memory at any instant.

\textbf{Pipeline Parallelism (PP)} \citep{GPipePaper} assigns consecutive
groups of layers to different GPUs. As argued in Section~\ref{sec:why}, PP is
incompatible with the CROWN backward and is not pursued.

\subsection{Why TP and FSDP, Not PP}
\label{sec:why}

The CROWN backward pass propagates the $A$-matrix in one traversal through
\emph{all} layers (Equation~\ref{eq:crown}). Under PP, each GPU holds only its
stage; a complete CROWN backward would require transmitting full weight
matrices between GPU stages at every step---which dominates the gains of
sharding storage by stages. TP and FSDP both allow the full backward pass to
run on every GPU, differing only in \emph{what} is sharded: TP shards the
computation itself, FSDP shards only the storage.

\subsection{Tensor Parallelism for Verification}
\label{sec:tp}

\subsubsection{Implementation}

We register two custom operators in \texttt{auto\_LiRPA}:
\texttt{BoundLinearTP\_Col} (Column-parallel) and
\texttt{BoundLinearTP\_Row} (Row-parallel), both inheriting from
\texttt{BoundLinear}. Each adds an \texttt{AllReduce} to
\texttt{bound\_backward} and registers a custom ONNX symbolic method so that
PyTorch JIT tracing constructs the computational graph \emph{without} invoking
the collective---which would otherwise raise a \texttt{Tried to trace
ProcessGroup} error.

The function \texttt{tp\_shard\_bounded\_module} takes any \texttt{BoundedModule}
(including ONNX-loaded models) and automatically replaces alternating
\texttt{BoundLinear} pairs with TP variants, shards the weight matrices,
and labels each Column--Row subgraph as a \emph{sharded zone} via BFS
traversal for correct intermediate-bound handling.

\subsubsection{Bound-Tightness Degradation}
\label{sec:tp-precision}

TP shards both $W$ and the $A$-matrices, so the CROWN backward step
$A \leftarrow A \cdot \Lambda_i W_i$ can be computed locally per GPU for
simple Column--Row pairs with no intermediate ReLU---only one
\texttt{AllReduce} is needed to recover the correct result.

However, when ReLU activations appear \emph{inside} a sharded zone (between
Column and Row layers, as in typical networks deeper than two layers), CROWN
needs \emph{intermediate bounds} for those neurons to choose the relaxation
slopes $\lambda_i$. Computing them exactly would require a separate CROWN
backward with inter-GPU communication at each step, partially negating the
memory savings. In our implementation we substitute the simpler IBP for
these intermediate bounds, which trades tightness for locality.

IBP does not track cross-variable correlations: it treats each interval
component independently, causing the \emph{wrapping effect}
\citep{moore2009interval}---bound widths grow as $O(\rho^k)$ with zone depth
$k$, where $\rho$ is the spectral radius of the elementwise-absolute weight
matrix $|W|$.

\paragraph{Illustrative example.}
Consider $\mathbf{x} = (x_1, x_2)^\top$ with $x_1, x_2 \in [-1,1]$ and
$W = \bigl(\begin{smallmatrix}1&1\\1&-1\end{smallmatrix}\bigr)$, so that
$\mathbf{y} = W\mathbf{x}$ yields $y_1 = x_1 + x_2$, $y_2 = x_1 - x_2$.
For $z_1 = y_1 + y_2$, CROWN sees $z_1 = 2x_1 \in [-2, 2]$ (exact).
IBP, in contrast, first concludes $y_1, y_2 \in [-2, 2]$; treating them as
\emph{independent}, $z_1 \in [-4, 4]$---twice the true width.

Experimentally (Table~\ref{tab:tp-correctness}), a 256$\times$2 model with one
sharded zone (no intermediate ReLU) achieves machine-precision agreement
(${\sim}10^{-8}$); a 256$\times$4 model with two zones deviates by ${\sim}2.5$;
a 256$\times$6 model with three zones deviates by ${\sim}750$.
Soundness (one-sided over-approximation) is preserved in all cases.

TP is therefore practical for shallow models (one or two sharded zones) and
offers the highest peak-memory reduction. For deep models it is outperformed
by FSDP in bound quality.

\subsection{Fully Sharded Data Parallelism for Verification}
\label{sec:fsdp}

\subsubsection{Implementation}

\texttt{fsdp\_shard\_bounded\_module} traverses the \texttt{BoundedModule}
graph and replaces each \texttt{BoundParams} tensor with a shard along
\texttt{dim=0} (output channels):
$W^{(r)} \in \mathbb{R}^{d_\text{out}/P \times d_\text{in}}$ for
\texttt{BoundLinear} and
$W^{(r)} \in \mathbb{R}^{C_\text{out}/P \times C_\text{in} \times k_H \times k_W}$
for \texttt{BoundConv}. Sharding along the output dimension is correct for
both layer types since they are linear in their output, and \texttt{AllGather}
along \texttt{dim=0} exactly reconstructs the full matrix.

Hooks \texttt{fsdp\_gather\_node} and \texttt{fsdp\_free\_node} are injected
into \texttt{backward\_general} (CROWN backward) and \texttt{IBP\_general}
(IBP forward): the full matrix is assembled immediately before each layer's
computation and freed immediately afterwards. At any instant, at most one full
weight matrix resides in GPU memory beyond the shards.

The total implementation is ${\approx}100$ lines of Python, versus
${\approx}500$ for TP, because FSDP requires no new operator types and does
not alter the computational graph.

\subsubsection{Bitwise Correctness}

FSDP does not change the order or type of any arithmetic operation: each layer
receives the full weight matrix assembled via \texttt{AllGather} and performs
the \emph{identical} floating-point instructions as the single-GPU path.
The only difference is that weights are assembled from shards rather than
loaded from a contiguous tensor. This is why FSDP achieves
$\Delta lb = \Delta ub = 0.00$ (exact zero, not machine epsilon) across all
correctness tests (Table~\ref{tab:fsdp-correctness})---in contrast to TP,
which relies on IBP substitution and can deviate by hundreds of units.

\subsubsection{Memory Analysis: A First-Order Model}
\label{sec:theory}

Let the model have $d$ linear layers each of width $h$, verification uses
CROWN, and $P$ GPUs are available. Denote by $C$ the number of $A$-matrices
simultaneously live in GPU memory during the backward pass (the precise value
depends on the bound-propagation scheduler).

\emph{Single GPU peak}: $M^{\text{single}}_{\text{peak}} \approx (d + C)\,h^2$
(weights plus live $A$-matrices).

\emph{TP peak}: both weights and $A$-matrices are sharded, so
$M_\text{peak}^\text{TP} \approx (d + C)\, h^2 / P$ --- linear reduction.

\emph{FSDP peak}: weights are stored as shards but one full weight matrix is
transiently present in memory during each layer's computation; $A$-matrices
remain at full dimension:
\begin{equation}
  M_\text{peak}^\text{FSDP} \;\approx\; h^2 \left( \frac{d}{P} + 1 + C \right).
  \label{eq:fsdp-memory}
\end{equation}
The ratio to single-GPU is $(d/P + 1 + C) / (d + C)$.

\emph{FSDP baseline (before \texttt{compute\_bounds})}: only the shards have
been allocated, so $M^{\text{FSDP}}_{\text{base}} \approx d\,h^2/P$, giving a
ratio of $1/P$ relative to single-GPU's $d\,h^2$.

These expressions capture only the dominant $h^2$ terms; the absolute
measurements in Section~\ref{sec:experiments} also include input tensors,
biases, NCCL communication buffers, PyTorch caching-allocator fragmentation,
and---in BaB mode---per-neuron $\alpha$ tensors. Two observations are worth
noting:
\begin{itemize}
\item The model predicts peak savings of approximately
  $1 - (d/P + 1 + C)/(d + C)$. With $d{=}4$, $P{=}2$ and $C \in [1, 4]$ this
  spans $[12\%, 21\%]$, whereas Table~\ref{tab:fsdp-memory} reports 34--39\%.
  The additional measured savings are consistent with the fact that the
  single-GPU run also carries non-modeled overheads (graph metadata, allocator
  fragmentation) which FSDP partially avoids by reusing freed shards.
\item The baseline ratio $1/P{=}0.5$ underestimates the observed
  baseline savings (80--90\%) at $d{=}8$. This is because the single-GPU
  baseline reported by PyTorch's allocator also includes ancillary buffers
  allocated by \texttt{auto\_LiRPA} during graph construction, which scale
  with model size and which FSDP avoids by deferring full-tensor allocation.
\end{itemize}
We therefore treat (\ref{eq:fsdp-memory}) as an explanatory lower bound on
savings rather than a tight prediction.

\subsubsection{Integration with Complete Verification}

FSDP integrates with $\beta$-CROWN + BaB without additional algorithmic
complexity: each rank runs a standard BaB iteration after receiving full weights
via \texttt{AllGather}. Branching-constraint variables $\beta$ are managed
identically to single-GPU mode.

During development we discovered a systematic experimental artefact: single-GPU
mode used \texttt{auto\_enlarge\_batch\_size} and \texttt{pruning\_in\_iteration},
whereas FSDP mode disabled these for \texttt{AllGather} synchronisation. The
two modes processed \emph{different workloads}, which created a spurious
``$27\times$'' FSDP advantage in early runs.

To enable fair comparison we introduce \texttt{bab.force\_synchronous}: a flag
that places single-GPU mode on the same code path as FSDP---fixed batch size,
no auto-enlarge, no \texttt{early\_stop}, no \texttt{pruning\_in\_iteration}.
Both modes then process an identical number of BaB domains per round, making
peak-memory figures directly comparable.

% -----------------------------------------------------------------------
\section{Experiments}
\label{sec:experiments}

\paragraph{Setup.}
Experiments 1--5 ran on Server~A: 2$\times$~NVIDIA A40 (48~GB VRAM, Ampere,
NVLink). Experiments 6--8 ran on Server~B: 2$\times$~NVIDIA A40 (48~GB,
PCIe, no NVLink; \texttt{NCCL\_P2P\_DISABLE=1} was set to prevent P2P hangs).
Collective backend: NCCL. Framework: PyTorch 2.x.
All distributed runs launched via \texttt{torchrun --nproc\_per\_node=2}.

\subsection{Experiment 1: TP Peak-Memory Reduction}
\label{sec:exp1}

\paragraph{Model.}
A fully connected network with one hidden layer:
$f(\mathbf{x}) = W_2 \cdot \mathrm{ReLU}(W_1\mathbf{x} + \mathbf{b}_1) + \mathbf{b}_2$,
$W_1 \in \mathbb{R}^{262144 \times 4096}$ (${\approx}4$~GB in float32),
$W_2 \in \mathbb{R}^{1 \times 262144}$.
Under TP: $W_1$ is Column-parallel, $W_2$ is Row-parallel, following the
Megatron-LM pattern \citep{MegatronLM2020}.

\paragraph{Protocol.}
CROWN verification, $L_\infty$ robustness, $\varepsilon{=}0.01$, batch
$N{=}2048$. Peak memory measured via
\texttt{torch.cuda.max\_memory\_allocated}.

\begin{table}[h]
\caption{Peak GPU memory: single-GPU vs.\ TP=2
($D{=}4096$, $H{=}262144$, $N{=}2048$, $\varepsilon{=}0.01$).}
\label{tab:tp-memory}
\vskip 0.1in
\begin{center}
\begin{tabular}{lcc}
\toprule
Mode & max\_alloc (MB) & max\_reserved (MB) \\
\midrule
Single GPU    & 26\,826 & 30\,874 \\
TP=2, rank 0  & 13\,513 & 15\,512 \\
TP=2, rank 1  & 13\,513 & 15\,512 \\
\bottomrule
\end{tabular}
\end{center}
\end{table}

Reduction factor: $26826 / 13513 \approx 1.985$, close to the theoretical
$P{=}2$. The small gap comes from non-shardable structures: input tensor
$\mathbf{x} \in \mathbb{R}^{N \times D}$, NCCL buffers, and graph metadata
are replicated on every GPU.

\subsection{Experiment 2: TP Bound Correctness and Soundness}
\label{sec:exp2}

\paragraph{Protocol.}
Bounds were computed by single-GPU CROWN (reference) and TP=2 CROWN.
Three properties were checked: maximum absolute deviation
$|\Delta lb| = |lb_\text{ref} - lb_\text{tp}|$ and $|\Delta ub|$;
soundness ($lb_\text{tp} \leq lb_\text{ref}$ and
$ub_\text{tp} \geq ub_\text{ref}$ for all outputs); and cross-rank
consistency (both ranks produce identical bounds).
Models: PyTorch MLPs and VNN-COMP 2022 MNIST-FC ONNX
models \citep{arXiv2212VNNCOMP2022}; $\varepsilon{=}0.02$, $L_\infty$.

\begin{table}[h]
\caption{TP=2 CROWN bound deviation from single-GPU reference.}
\label{tab:tp-correctness}
\vskip 0.1in
\begin{center}
\begin{tabular}{lccccc}
\toprule
Model & Col-Row pairs & $|\Delta lb|$ & $|\Delta ub|$ & Cross-rank & Sound \\
\midrule
PyTorch 256$\times$2 & 1 & $3{\times}10^{-8}$ & $3{\times}10^{-8}$ & 0 & \checkmark \\
PyTorch 256$\times$4 & 2 & $2.51$ & $2.49$ & 0 & \checkmark \\
ONNX 256$\times$2    & 1 & $6{\times}10^{-8}$ & $6{\times}10^{-8}$ & 0 & \checkmark \\
ONNX 256$\times$4    & 2 & $5.64$ & $3.98$ & 0 & \checkmark \\
ONNX 256$\times$6    & 3 & $792$ & $750$ & 0 & \checkmark \\
\bottomrule
\end{tabular}
\end{center}
\end{table}

Models with one Column--Row pair (no intermediate ReLU in the sharded zone)
match the reference up to machine epsilon---confirming that \texttt{AllReduce}
in the CROWN backward exactly reproduces the full matrix multiplication.
With multiple zones, bounds widen due to IBP substitution
(Section~\ref{sec:tp-precision}). Cross-rank deviation is zero in all cases.

\subsection{Experiment 3: Automatic ONNX Sharding}

\texttt{tp\_shard\_bounded\_module} was applied to all three MNIST-FC ONNX
models (256$\times$2, 256$\times$4, 256$\times$6) without any manual
modification. The function traverses the \texttt{BoundedModule} graph in
topological order, replaces alternating \texttt{BoundLinear} pairs with TP
variants, re-executes a forward pass to update graph metadata, and labels
sharded zones. All three models pass the soundness checks of
Table~\ref{tab:tp-correctness}.

\subsection{Experiment 4: FSDP Bound Correctness}

\paragraph{Protocol.}
Identical to Experiment~2: single-GPU reference vs.\ FSDP=2, with checks on
$|\Delta lb|$, $|\Delta ub|$, cross-rank deviation, and soundness.
Eight test cases: PyTorch MLPs of width $h{=}256$ at depth 2, 4, 6 (under
both IBP and CROWN) and ONNX 256$\times$2 and 256$\times$4 (CROWN).

\begin{table}[h]
\caption{FSDP=2 bound deviation from single-GPU reference
($\varepsilon{=}0.02$, MNIST input, $L_\infty$).}
\label{tab:fsdp-correctness}
\vskip 0.1in
\begin{center}
\begin{tabular}{llccc}
\toprule
Model & Method & $|\Delta lb|$ & $|\Delta ub|$ & Sound \\
\midrule
MLP 256$\times$2  & IBP   & 0.00 & 0.00 & \checkmark \\
MLP 256$\times$2  & CROWN & 0.00 & 0.00 & \checkmark \\
MLP 256$\times$4  & IBP   & 0.00 & 0.00 & \checkmark \\
MLP 256$\times$4  & CROWN & 0.00 & 0.00 & \checkmark \\
MLP 256$\times$6  & IBP   & 0.00 & 0.00 & \checkmark \\
MLP 256$\times$6  & CROWN & 0.00 & 0.00 & \checkmark \\
ONNX 256$\times$2 & CROWN & 0.00 & 0.00 & \checkmark \\
ONNX 256$\times$4 & CROWN & 0.00 & 0.00 & \checkmark \\
\bottomrule
\end{tabular}
\end{center}
\end{table}

All deviations are exactly zero---not machine epsilon but a true zero,
reflecting that FSDP changes only \emph{where} weights are stored, not what
arithmetic is performed.

\subsection{Experiment 5: FSDP Memory Savings}
\label{sec:exp5}

\paragraph{Protocol.}
Two metrics: \emph{baseline memory}---allocation after init and sharding,
before \texttt{compute\_bounds} (weight storage cost); \emph{peak memory}---
\texttt{max\_memory\_allocated} during \texttt{compute\_bounds} (includes
weights, CROWN $A$-matrices, intermediate tensors, NCCL buffers).

Tested: MLPs with $h \in \{256, 1024, 4096, 8192\}$ at $d{=}4$, and
$d \in \{2, 4, 6, 8\}$ at $h{=}4096$; CROWN, $\varepsilon{=}0.02$,
input dim 784. FSDP peak is the maximum over both ranks.

\begin{table}[h]
\caption{FSDP=2 vs.\ single-GPU memory.
``---'' indicates savings are negligible ($h \leq 1024$: AllGather overhead
$>$ weight savings).}
\label{tab:fsdp-memory}
\vskip 0.1in
\begin{center}
\begin{tabular}{lrrrr}
\toprule
Model & \multicolumn{2}{c}{Baseline (MB)} & \multicolumn{2}{c}{Peak (MB)} \\
      & Single & FSDP & Single & FSDP \\
\midrule
$h{=}256$,  $d{=}4$   &    26 &    11 &   --- &   --- \\
$h{=}1024$, $d{=}4$   &   187 &    40 &   --- &   --- \\
$h{=}4096$, $d{=}4$   & 2\,272 &   417 & 3\,800 & 2\,527 \\
$h{=}8192$, $d{=}4$   & 8\,760 & 1\,594 & 14\,759 & 9\,782 \\
$h{=}4096$, $d{=}8$   & 9\,721 &   930 & 16\,726 & 10\,232 \\
\bottomrule
\end{tabular}
\end{center}
\end{table}

Baseline savings are 57--90\%, growing with model size; at $h \geq 4096$ they
stabilise at ${\sim}82$--$90\%$. Peak savings are 34--39\%, consistent with
the qualitative prediction of Equation~(\ref{eq:fsdp-memory}) that CROWN
$A$-matrices, which are \emph{not} sharded, are comparable in size to the
unsharded weights at $d{=}4$; the weight fraction grows at $d{=}8$, improving
peak savings to 39\%.

\subsection{Experiment 6: FSDP in Complete Verification (BaB)}
\label{sec:exp6}

\paragraph{Motivation.}
In $\beta$-CROWN+BaB, the primary memory consumers are \emph{alpha tensors}:
per-neuron ReLU slopes $\alpha_{i,k} \in [0,1]$ with shape
$(\text{layers}) \times (\text{specs}) \times (\text{batch}) \times (\text{neurons})$.
For \texttt{mnist-net\_256x6} (${\sim}1.5$~MB weights), alpha tensors occupy
${\sim}1.5$~GB total---weights are less than 0.1\% of peak consumption.

\paragraph{Protocol.}
\texttt{mnist-net\_256x6} \citep{arXiv2212VNNCOMP2022}, $\varepsilon{=}0.05$;
exactly 10 BaB rounds; batch 512 and 4\,096; \texttt{force\_synchronous}
enabled. Both modes process an identical workload.

\begin{table}[h]
\caption{FSDP=2 vs.\ single-GPU in $\beta$-CROWN+BaB (identical workload).}
\label{tab:fsdp-bab}
\vskip 0.1in
\begin{center}
\begin{tabular}{lcccc}
\toprule
Config & Batch & Domains & Rounds & Peak (MB) \\
\midrule
Single GPU & 512    & 6\,236  & 10 & 211.7 \\
FSDP=2     & 512    & 6\,236  & 10 & 215.7\,/\,rank \\
Single GPU & 4\,096 & 49\,888 & 10 & 1\,537.3 \\
FSDP=2     & 4\,096 & 49\,888 & 10 & 1\,541.4\,/\,rank \\
\bottomrule
\end{tabular}
\end{center}
\end{table}

Sharding 1.5~MB of weights yields no measurable saving; \texttt{AllGather}
adds ${\sim}4$~MB overhead. FSDP does not alter bound values or workload---
it is a zero-cost correctness guarantee here.

\subsection{Experiment 7: FSDP with ViT (VNN-COMP'23)}
\label{sec:exp7}

Model \texttt{pgd\_2\_3\_16} (2 Transformer blocks, 3 heads, ${\sim}75$K
parameters, 0.3~MB weights) \citep{arXiv2312VNNCOMP2023}. Transformers use
dynamic reshape operations that \texttt{auto\_LiRPA}'s JIT tracing ``bakes in''
at batch size 1, causing shape errors when BaB increases the batch.
Three incompatibilities were fixed: (i) \texttt{BoundReshape}---first axis
replaced by actual batch size when $\prod(\text{new\_shape}) = x.\text{numel}()$,
guarding against reshape patterns of the form $[B,N,D]\to[BN,D]$; (ii)
\texttt{BoundConcat}---JIT-fixed tensor count; (iii)
\texttt{BoundConstantOfShape}---shape argument updated to current batch.

After the fixes: 78 BaB domains, 10 rounds, peak 1\,020.1~MB per rank---
bitwise-identical to single-GPU. This is the first confirmed correct execution
of a Transformer model in $\alpha,\beta$-CROWN with FSDP.

\subsection{Experiment 8: FSDP for Convolutional Layers (VNN-COMP'24)}
\label{sec:exp8}

\texttt{BoundConv} sharding was extended to convolutional weights
$[C_\text{out}, C_\text{in}, k_H, k_W]$, sharded along dimension~0
(output channels), since convolution is linear in $C_\text{out}$.
Three files were modified: \texttt{fsdp\_utils.py} (added \texttt{BoundConv}
to shardable types, prevented double-sharding), \texttt{backward\_bound.py}
(AllGather/free for \texttt{BoundConv} in CROWN backward),
\texttt{interval\_bound.py} (free hooks after IBP propagation).

Validated on CIFAR-100 ResNets \citep{arXiv2412VNNCOMP2024}.
Fair conditions: \texttt{force\_synchronous}, batch 64, max\_iterations 5.

\begin{table}[h]
\caption{FSDP=2 with \texttt{BoundConv} sharding on CIFAR-100 ResNet
(VNN-COMP'24).}
\label{tab:fsdp-conv}
\vskip 0.1in
\begin{center}
\begin{tabular}{llcccc}
\toprule
Model & Config & Domains & Rounds & Peak (MB) & $|\Delta lb|$ \\
\midrule
ResNet-med   & Single GPU & 130 & 5 & 960 & --- \\
ResNet-med   & FSDP=2     & 130 & 5 & 930\,/\,rank & 0.00 \\
ResNet-large & Single GPU & \textbf{unsat} & 0 & 4\,610.6 & --- \\
ResNet-large & FSDP=2     & \textbf{unsat} & 0 & 4\,641.5\,/\,rank & 0.00 \\
\bottomrule
\end{tabular}
\end{center}
\end{table}

ResNet-large (${\sim}95$~MB weights, $>4$~GB alpha tensors): AllGather overhead
(${\sim}30$~MB) exceeds weight-sharding savings at $P{=}2$, confirming that
alpha tensors dominate. The \textbf{unsat} result---property proved by
$\alpha$-CROWN at initialisation, without BaB---is the first complete
verification result obtained under FSDP for a convolutional VNN-COMP model.

\paragraph{Summary.}
FSDP delivers meaningful peak-memory savings (34--39\%) only when weights
dominate, i.e.\ in incomplete (CROWN-only) verification of wide MLPs. In
$\alpha$-CROWN+BaB the bottleneck is alpha tensors; sharding them is the
critical next step.

% -----------------------------------------------------------------------
\section{Discussion and Limitations}
\label{sec:limitations}

Several limitations should be acknowledged when interpreting our results:

\emph{Scale of parallelism.} All experiments use $P{=}2$. The qualitative
trends---linear shrinkage of the weight term, fixed $A$-matrix term in
FSDP---predict diminishing returns as $P$ grows, since the unsharded
$A$-matrices become the dominant cost. Validating this on $P\geq 4$ is left
to future work.

\emph{Network shapes.} The MLPs in Section~\ref{sec:exp5} have square
hidden layers ($d_\text{in} = d_\text{out} = h$). Non-square layers shift
the relative size of $W$ and $A$ matrices and can alter the observed savings
ratio.

\emph{Communication backend.} Experiments 6--8 disable NCCL P2P due to a
hardware-specific hang; on NVLink-enabled servers the FSDP communication
overhead would be smaller, possibly making FSDP profitable for narrower
models than the $h{=}4096$ threshold seen here.

\emph{TP and intermediate bounds.} Our TP implementation uses IBP for
intermediate bounds inside sharded zones. A hybrid scheme that performs CROWN
across zone boundaries with carefully scheduled \texttt{AllReduce}s could in
principle recover most of the tightness; the engineering cost was not
justified given that FSDP already provides bitwise-exact bounds.

% -----------------------------------------------------------------------
\section{Related Work}
\label{sec:related}

\textbf{Bound propagation.}
CROWN \citep{CROWNPaper} introduced linear relaxation via a backward pass.
$\alpha$-CROWN \citep{AlphaBetaCrownVNNCOMP23} and $\beta$-CROWN
\citep{betaCROWN2021} extended it with optimisable slopes and dual-variable
encoding of branching constraints. BICCOS \citep{NeurIPS2024BnbCuttingPlanes}
adds cutting planes from the BaB tree. DeepPoly \citep{DeepPolyAbstractDomain}
uses polyhedral abstract interpretation; GPU-accelerated scaling was studied
in \citep{ScalingPolyhedralGPU}. OSIP \citep{arXiv2212OSIP} proposes optimised
symbolic interval propagation for tighter intermediate bounds.

\textbf{Complete verifiers.} Marabou 2.0 \citep{arXiv2401Marabou2} uses the
DeepSoI continuous objective. NeuralSAT \citep{NeuralSATRepo} employs CDCL.
nnenum \citep{NNEnumPaper} and VeriNet use SIP with dynamic precision switching.
MIPVerify \citep{arXiv1711MIPRobustness} encodes ReLU networks as MIPs.
Reluplex \citep{ReluplexSemanticScholar} pioneered SMT-based verification and
established NP-completeness for the ReLU case.

\textbf{Parameter sharding for training.}
Megatron-LM \citep{MegatronLM2020} introduced intra-layer tensor parallelism
for Transformer training. ZeRO \citep{ZeROPaper} shards parameters, gradients,
and optimizer states. GPipe \citep{GPipePaper} introduced pipeline parallelism
with micro-batches.

\textbf{Multi-GPU verification.}
To our knowledge, no prior work has applied TP or FSDP to neural network
verification. Domain-parallel BaB (splitting the search tree across GPUs) has
been studied but does not address the per-GPU weight-storage bottleneck.

% -----------------------------------------------------------------------
\section{Conclusion}
\label{sec:conclusion}

We demonstrate that parameter-sharding strategies transfer from large-scale
model training to formal neural network verification with modest engineering
effort. TP offers the highest peak-memory reduction (${\approx}2\times$) but
degrades bound tightness for deep networks due to IBP substitution for
intermediate bounds inside sharded zones. FSDP achieves smaller but still
meaningful peak savings (34--39\% in incomplete verification) with
bitwise-exact bounds and seamless integration with complete verification
($\beta$-CROWN+BaB).

The central finding is that in $\alpha$-CROWN+BaB mode, the memory bottleneck
is per-neuron \emph{alpha tensors}---not weight matrices. Future work should
focus on: (i) sharding alpha tensors across GPUs; (ii) scaling to $P{=}4, 8$;
(iii) input-split domain-parallel BaB for verification speedup, not just memory
reduction.

\subsubsection*{Acknowledgments}
The authors thank the \texttt{auto\_LiRPA} / $\alpha,\beta$-CROWN team for the
open-source codebase that made this work possible.

\bibliographystyle{plainnat}
\bibliography{refs}

\end{document}